%% file: main.tex
\documentclass[runningheads,a4paper]{llncs}

\usepackage[english]{babel}
\usepackage[utf8x]{inputenc}
\usepackage{amsmath}
\usepackage{amssymb}
\usepackage{graphicx}
\usepackage[colorinlistoftodos]{todonotes}
\usepackage{url}
\usepackage{graphicx}
\usepackage{array}
\usepackage{multirow}
\usepackage{subfigure}
\usepackage{afterpage}
\usepackage{color, ifpdf}
\usepackage{bm}
\usepackage{wrapfig}
\usepackage{soul} 

\newcolumntype{L}[1]{>{\raggedright\let\newline\\\arraybackslash\hspace{0pt}}m{#1}}
\newcolumntype{C}[1]{>{\centering\let\newline\\\arraybackslash\hspace{0pt}}m{#1}}
\newcolumntype{R}[1]{>{\raggedleft\let\newline\\\arraybackslash\hspace{0pt}}m{#1}}

\DeclareMathOperator*{\argmin}{arg\,min}

\newcommand{\carlos}{\color{blue}}

\begin{document}

\title{Towards a glaucoma risk index based on simulated hemodynamics from fundus images}
\titlerunning{Simulation of the retinal hemodynamics for glaucoma assessment}  

\author{Jos\'e~Ignacio~Orlando\inst{1} \and Jo\~ao~Barbosa~Breda\inst{2} \and Karel~van~Keer\inst{2} \and Matthew~B.~Blaschko\inst{3} \and Pablo~J.~Blanco\inst{4} \and Carlos~A.~Bulant\inst{1}}


\authorrunning{Orlando \textit{et al.}} 


\tocauthor{Jos\'e~Ignacio~Orlando, Jo\~ao~Barbosa~Breda, Karel~van~Keer, Matthew~B.~Blaschko, Pablo~Javier~Blanco and Carlos~A.~Bulant}


\institute{CONICET - Pladema Institute, UNICEN, Argentina 
\and 
Research Group Ophthalmology, KU Leuven, Leuven, Belgium 
\and
ESAT-PSI, KU Leuven, Leuven, Belgium
\and
National Laboratory for Scientific Computing, LNCC/MCTIC, Petr\'opolis, Brazil.
}

\maketitle

\input{abstract.tex}

\input{introduction.tex}
\input{methods.tex}

\input{results.tex}

\input{discussion.tex}

\noindent\\
\textbf{Acknowledgements}\\
This work is funded by ANPCyT PICTs 2016-0116 and start-up 2015-0006, the 
FWO 
G0A2716N, WWTF VRG12-009 and a NVIDIA Hardware Grant. JIO is now with OPTIMA, Medical University of Vienna, Austria.

\bibliography{RetinalImagingReport.bib}
\bibliographystyle{splncs}

\end{document}

%% file: abstract.tex
\begin{abstract}
Glaucoma is the leading cause of irreversible but preventable blindness in the world. Its major treatable risk factor is the intra-ocular pressure, although other biomarkers are being explored to improve the understanding of the pathophysiology of the disease. It has been recently observed that glaucoma induces changes in the ocular hemodynamics. However, its effects on the functional behavior of the retinal arterioles have not been studied yet. In this paper we propose a first approach for characterizing those changes using computational hemodynamics. The retinal blood flow is simulated using a 0D model for a steady, incompressible non Newtonian fluid in rigid domains. The simulation is performed on patient-specific arterial trees extracted from fundus images. We also propose a novel feature representation technique to comprise the outcomes of the simulation stage into a fixed length feature vector that can be used for classification studies. Our experiments on a new database of fundus images show that our approach is able to capture representative changes in the hemodynamics of glaucomatous patients. Code and data are publicly available in \url{https://ignaciorlando.github.io}.
\end{abstract}

%% file: introduction.tex
\section{Introduction}
\label{sec:introduction}

Glaucoma is a neurodegenerative condition that is the the leading cause of irreversible but preventable blindness~\cite{tham2014global}.
The disease is characterized by the interruption in the communication between the retinal photoreceptors and the brain, preventing visual signals to be carried to the brain via the optic nerve~\cite{tham2014global}. These progressive changes are asymptomatic until advanced stages of the disease, when undiagnosed patients notice the disease when a considerable amount of visual field has been lost. 
The major treatable glaucoma risk factor is the intra-ocular pressure (IOP), although
individuals with high IOP might not develop glaucoma, and subjects with normal IOP values can develop it~\cite{Harris_ImportanceOfMathematicalModelInGlaucoma-2013}. 
Hence, alternative genetic, vascular, anatomical or systemic factors are being extensively studied 
to better understand the pathophysiology of glaucoma~\cite{Harris_ImportanceOfMathematicalModelInGlaucoma-2013,barbosa2018clinical}. 

The ocular hemodynamics has long been observed to be affected by glaucoma~\cite{Harris_ImportanceOfMathematicalModelInGlaucoma-2013,abegao2016ocular}
through imaging modalities such as color Doppler imaging or laser Doppler flowmetry, among others~\cite{abegao2016ocular}.
These tools provide non-invasive measurements of hemodynamic parameters from ocular vascular structures such as the central retinal artery (CRA), the ophthalmic artery (OA) or the retinal capillary network. 
In turn, the analysis of the hemodynamic characteristics of the retinal arterioles and venules, easily accessible through fundus photography, has been limited despite the availability of this imaging modality for several decades. 

\begin{figure}[t!]
\centering
\includegraphics[width=\textwidth]{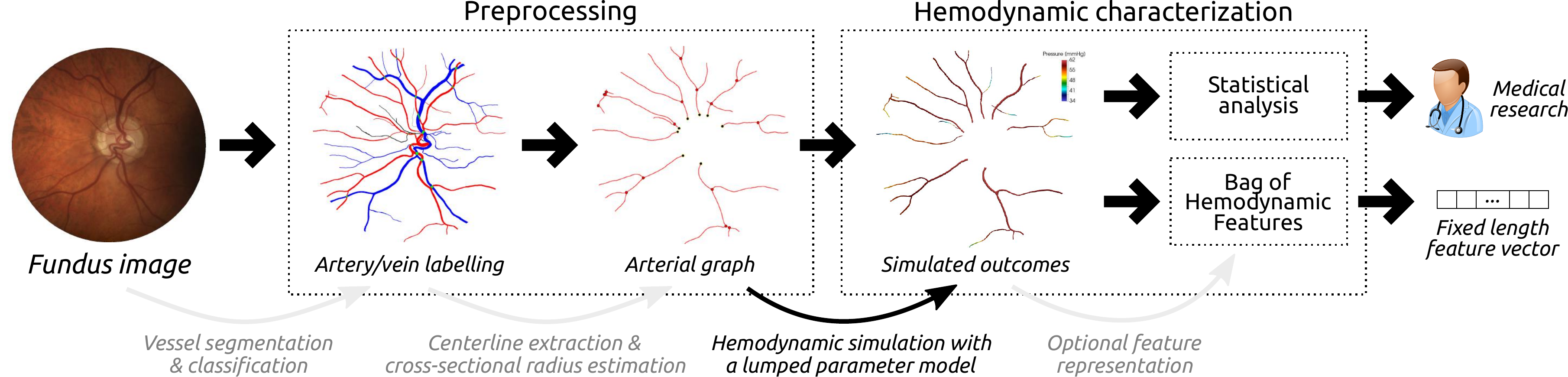}
\caption{Proposed method for characterizing patient-specific retinal hemodynamics.}
\label{fig:pipeline}
\end{figure}

In this paper, we propose a first approach to characterize the hemodynamics of the retinal microvasculature of glaucomatous and control subjects based on computer simulations performed on vascular networks extracted from fundus images (Fig.~\ref{fig:pipeline}). 
Our method is based on patient-specific graph representations of the arterial network that are used as the topological vascular substrate to build 0D models, which account for the steady, incompressible flow of a non Newtonian fluid (the blood) in rigid domains.
Although simpler than fully 3D~\cite{Lu_RCFD_DiabeticParafovealStudy-2016} or 2D~\cite{Liu_RCFD_FirstWork-2009} blood flow models, 0D models (also known as lumped parameter models) are computationally cheaper and, thus, enable the simulation of blood flow in large networks of vessels~\cite{Ganesan_RCFD_GraphArteryVeinsRats2-2010}.
Similar modeling approaches have been used before in the context of retinal hemodynamics simulation. In~\cite{Ganesan_RCFD_GraphArteryVeinsRats2-2010}, the blood hematocrit is incorporated as transport species in the model, considering constant cross-sectional radius per vessel segment in a network of arteries and veins of a mouse model.
More recently, 0D modeling has been applied in~\cite{Caliva_RCFD_DiabeticRetinopatyProgression-2017} for characterizing abnormal changes due to 
diabetic retinopathy. 
Although the simulation approach is similar to ours, a Newtonian fluid model was used in~\cite{Caliva_RCFD_DiabeticRetinopatyProgression-2017}. Furthermore, instead of using the whole retinal network, the numerical experiments have been performed in a single vascular bifurcation, which was manually isolated for each image. 
As a further novel aspect of the present study, the simulation outcomes are summarized in a fixed-length feature vector using a novel technique inspired in the bag of words~\cite{fei2005bayesian} technique for feature representation.
This allows the characterization approach to be applied in combination with linear classifiers to study its relationship with the presence/absence of the disease. We empirically validate this approach on a new database, namely LES-AV, comprising 22 fundus photographs with available manual segmentations of the retinal vessels and their expert classification into arteries and veins.

In summary, the key contributions of this study are: 
(i) a computational workflow for quantifying patient-specific hemodynamics non-invasively;
(ii) a feature representation technique to comprise the simulated outcomes into a fixed length feature vector; 
(iii) a first pilot study showing that glaucomatous patients exhibit changes in the hemodynamic variables with respect to control subjects; and
(iv) a new database of fundus images that is released, jointly with code and implementation details, in \url{https://ignaciorlando.github.io.}


%% file: methods.tex
\section{Materials and Methods}
\label{sec:methods}


The experiments were conducted on a new database of color fundus images, called LES-AV, comprising 22 images/patients with resolutions of $30^\circ$ FOV and $1444 \times 1620$ pixels (21 images), and $45^\circ$ FOV and $1958 \times 2196$ pixels (one image), with each pixel $=$ 6 $\mu m$. 
Demographic data about the groups are provided in Section~\ref{sec:results} 
and the supplementary materials.


\subsection{Preprocessing}
\label{subsec:preprocessing}

Patient-specific segmentations of the retinal vasculature must be first retrieved from a fundus image, and subsequently classified into arteries and veins. Although several methods have been introduced for this purpose~\cite{moccia2018blood}, we follow a semiautomatic approach to ensure a proper input for the simulation step. Hence, a first coarse segmentation of the vessels is obtained using a fully convolutional neural network~\cite{giancardo2017representation} (see supplementary materials for details). The segmentation is then manually refined to improve the vessel profiles and their connectivity, and each structure is manually labelled as artery or vein. Subsequently, the arteries are taken and their corresponding centerlines are automatically identified using a skeletonization method~\cite{rumpf2002continuous}. The result is a binary mask $T$ with $N_I$ connected components. Since the vasculature is observed as a 2D projection of the original 3D structure in a fundus image, the vessels in the optic disc (OD) are generally overlapped with each other, making unfeasible the identification of the roots of the trees in any realistic scenario. To mitigate this issue, the arterial segments inside the OD are removed from $T$, and a root pixel $I_t$ is automatically chosen for each connected component $T_t \in T$, such that $I_t$ corresponds to the closest pixel to the OD center. Subsequently, each tree $T_t$ is modeled as a graph structure $G_t = \langle I_t, \mathcal{V}_t, \mathcal{E}_t \rangle $, where each node $v_j \in \mathcal{V}_t$ is a set of branching pixels in $T_t$, and each edge $e_{(i,j)} \in \mathcal{E}_t$ is a segment of centerline pixels connecting two branching points $v_i$ and $v_j$. The pixels in the skeletonized arterial trees are automatically classified as part of a node or an edge in $G_t$ by analyzing their neighborhood.
Finally, the cross-sectional lumen radius at each centerline pixel is estimated from the segmentation using the Euclidean transform~\cite{Maurer_EuclideanTransformUsedByMATLAB-2003}.

\subsection{Simulation of the retinal hemodynamics}
\label{subsec:simulation}

The simulation is performed on a patient-specific computational mesh associated to the graphs $G_t$ (Section~\ref{subsec:preprocessing}). This structure comprises the key information required by the model to perform the simulation: (a) the topological connectivity of the arterial segments, through their branching points; (b) the pixel-wise information of the arterial radius, $r_i$, $i=\{1,...,N\}$, with $N$ the number of centerline pixels (a computational node per pixel is considered); (c) the root points, which are the network inlets $I_t$, $t=1,...,N_I$; and the network outlets $O_m$, $m=1,...,N_O$.

A steady state 0D model 
is used to describe the incompressible flow of a non-Newtonian fluid in non-compliant vessels. 
Such a model is ideal for this scenario, as it allows to efficiently analyze different simulations, and provides physiologically reasonable coarse descriptions of the hemodynamics in vascular networks. 
The non Newtonian properties of the blood are incorporated 
by expressing the viscosity as a function of the vessel cross-sectional radius, to describe the Fahraeus-Lindqvist effect~\cite{Pries_VariablViscosity-1996}. 
As a result of the simulation process, the blood pressure ($P_i$) and the blood flow rate $Q_i$  are computed at each computational node ($i$).
The governing 0D equations are detailed below. 

The standard mass conservation and pressure continuity junction model is used at each bifurcation point, where node $i$ branches to nodes $j$ and $k$ (Eq.~\eqref{EQ::0DMmodel::Junctions}).

Given an arterial segment of $M$ pixels length, $M-1$ lumped parameter elements are used, where the mass conservation and the hydraulic analogue of the Ohm's law are considered. Then, for an element formed by nodes $i$ and $i+1$ (Eq.~\eqref{EQ::0DMmodel::Segments}),
the resistance to the flow at each segment is modeled with the Poiseuille's law: 
$R_{i,i+1} = 8\mu L / \pi r_{i,i+1}^4$
where $L$ is the segment length, $r_{i,i+1}=(r_i+r_{i+1})/2$ is the average radius and $\mu=\mu(r_{i,i+1})$ is the blood viscosity, which is assumed to be a function of the arterial radius (since $r<150 \mu$m), see the supplementary materials.

\begin{minipage}{0.40\linewidth}
\begin{equation}
\begin{cases}
Q_i &= Q_j + Q_k,\\
P_i &= P_j = P_k.
\end{cases}  
\label{EQ::0DMmodel::Junctions}
\end{equation}
\end{minipage}
\hspace{0.05\linewidth} 
\begin{minipage}{0.45\linewidth}
\begin{equation}
\begin{cases}
Q_{i+1} &= Q_i, \\
Q_{i+1} &= \dfrac{P_i - P_{i+1}}{R_{i,i+1}}.
\end{cases}  
\label{EQ::0DMmodel::Segments}
\end{equation}
\end{minipage}

Regarding boundary conditions, the pressure and flow rates, are prescribed at the inlet and outlets, respectively:
\begin{equation}
\begin{cases}
P_{I_l} = P_0, & \forall l=1,\ldots,N_I, \\
Q_{O_m} = \beta r_{O_m}^\gamma, & \forall m=1,\ldots,N_O.
\end{cases}  
\label{EQ::0DMmodel::BC} 
\end{equation}
The pressure at all the inlets is set to the mean CRA pressure, $P_0$, while the flow at the outlets is set to the value given by Murray's law
, which relates the flow rate to the outlet radius to the power of $\gamma$. Specifically, we set $\gamma=2.66$ \cite{Blanco_CCO-2013}, and putting $\beta = Q_T / \sum\limits_m r^\gamma_{O_m}$ ensures the total retinal flow $Q_T$.

The inlet pressure is set to the mean value of normotensive patients (i.e. $P_0=62.22$ mmHg).
Due to the lack of consensus on normal values for $Q_T$ ~\cite{pournaras2013retinal}, we proposed three scenarios, namely $Q_T^{\text{SC1}}=30, Q_T^{\text{SC2}}=45.6, Q_T^{\text{SC3}}=80 \quad \mu$l/m, which represent the mean (SC2) and extreme values (SC1 and SC3) for healthy subjects.
These values are preset for any given subject, without using patient-specific parameters. Further details about the physiological considerations of the parameters choice are provided in the supplementary materials.

The resulting real linear system of equations is solved using \texttt{LAPACK dgles} function through a QR factorization within a custom \texttt{C++} source code. 
After solving the system, for each centerline pixel $i$ in the network $G$, a feature vector $\mathbf{F}_i$ is computed, comprising: the flow rate $Q_i$, the blood pressure $P_i$, the mean cross sectional velocity $v_i=Q_i/(\pi r_i^2)$, the Poiseuille's resistance $R_i$, the Reynolds number $\text{Re}_i=\rho 2 r_i v_i/\mu_i$ and the wall shear stress WSS$_i=4\mu_i Q_i / (\pi r_i^3)$. Here, $\rho=1.040$ g/cm$^3$ is the blood density. 
At bifurcation points, the variables assume the value of the last computational node of the parent artery.


\subsection{Bag of Hemodynamic Features (BoHF)}
\label{subsec:bag-of-hemodynamic-features}

One key aspect of the functional characterization is how to construct a feature vector $\mathbf{F}$ for discriminating control subjects from diseased. As $\mathbf{F}_i$ is given for every centerline pixel {\carlos $i$}, it is necessary to find an alternative feature representation with a fixed length, suitable e.g.\ for linear discriminant analysis. 
To this end, we propose a new technique inspired by the Bag of Words (BoW) method~\cite{fei2005bayesian}, named as Bag of Hemodynamic Features (BoHF).

Our analysis is focused on three relevant structures for the graph $G^{(u)}$ of the $u$-th subject: the terminal and the bifurcation points, and the arterial segments. For every pixel $i$ being a terminal or a bifurcation point, the vector $\mathbf{F}_i$ is used. However, for a given arterial segment $s \in \mathcal{E}$, the average vector $\tilde{\mathbf{F}}_s$ over the pixels enclosed by the segment is used as a summary statistic. This allows our representation to lower its variance and to normalize for the length of an arterial segment. 
The resulting vectors associated to each arterial segment, bifurcation and terminal points are collected into a set, denoted $X$.

Let $S = \{ (X^{(u)}, y^{(u)}) \}, u \in \{1, ..., n\}$ be a training set of $n$ samples, where $X^{(u)}$ is the set of summary statistics at arterial segments, bifurcation and terminal points described above, and $y^{(u)} \in \mathcal{L} = \{ -1, +1 \}$ is its associated binary label (e.g. healthy or glaucomatous). Since the size of $X^{(u)}$ varies from one subject to another, our purpose is to map it into a feature vector $\mathbf{x}^{(u)} \in \mathbb{R}^d$, with $d$ fixed. In the traditional BoW approach, the $\mathbf{x}$ vectors are histograms of \textit{codewords} (analogous to words in text documents), and are obtained based on a \textit{codebook} $\mathcal{C}$ (analogous to a word dictionary). $\mathcal{C}$ must be designed such that it summarizes the most representative characteristics of each class in $\mathcal{L}$. Let us denote by $S_{-1}$ and $S_{+1}$ the set of negative and positive samples in $S$, respectively. A codebook $\mathcal{C} = \{ \mathcal{C}_{-1}, \mathcal{C}_{+1} \}$ can be automatically learned by applying $k$-means clustering in the $p$-dimensional feature space of each subset $S_{(\cdot)}$. The centroids of each of the learned $k$ clusters can be then taken as the codewords, resulting in a set of $k$ codewords for each class ($|\mathcal{C}| = d = 2k$).
Hence, the feature vector $\mathbf{x}$ can be obtained such that the $j$-th position in $\mathbf{x}$ corresponds to the number of nearest neighbors in $X^{(u)}$ to the $j$-th code in $\mathcal{C}$. If the codewords are representative of each label group, it is expected that the samples belonging to the positive/negative class will have higher/lower values in the last $k$ positions of $\mathbf{x}$.

Finally, the discrimination of glaucomatous patients from control subjects can be posed as a binary classification problem that can be tackled using any binary classification algorithm, e.g. $\ell_2$ regularized logistic regression. The linear discriminant function of the logistic regression model $f(\mathbf{x}) = \langle \beta, \mathbf{x} \rangle$ with parameters $\beta$ is obtained by solving:
$\hat{\beta} = \argmin_\beta \bigg[\lambda \|\beta\|_2^2 + \sum_{u=0}^{n} \log(1 + e^{-y^{(u)} \langle \beta, \mathbf{x}^{(u)} \rangle})\bigg]$. 




%% file: results.tex
\section{Results}
\label{sec:results}

The computational simulations of all scenarios resulted in mean values of pressure drop ($\Delta P$ to inlet, $P_0$) and blood velocity ($v$) consistent with previous studies in the human retinal circulation~\cite{Liu_RCFD_FirstWork-2009}.


The ability of the BoHF to capture the hemodynamic changes in the glaucomatous (G) from normal (H) subjects was empirically assessed using $\ell_2$ regularized logistic regression on LES-AV, in a leave-one-out cross-validation setting. The SC2 simulations were used for all the cases to avoid any bias in the classification performance. The values of $k \in \{2, ..., 15\}$ (BoHF length) and $\lambda \in 10^i, i \in \{-3, ..., 0, ..., 5\}$ (regularization parameter) were fixed to the values that maximized the accuracy in a held-out validation set randomly sampled from the training set. The features were centered to zero mean and unit variance using the training sets mean and standard deviation. The resulting area under the ROC curve was 0.70248, with an accuracy of 68.18\%.

\begin{figure}[t]
  \centering
  \subfigure[Samples]{\includegraphics[width=0.35\textwidth]{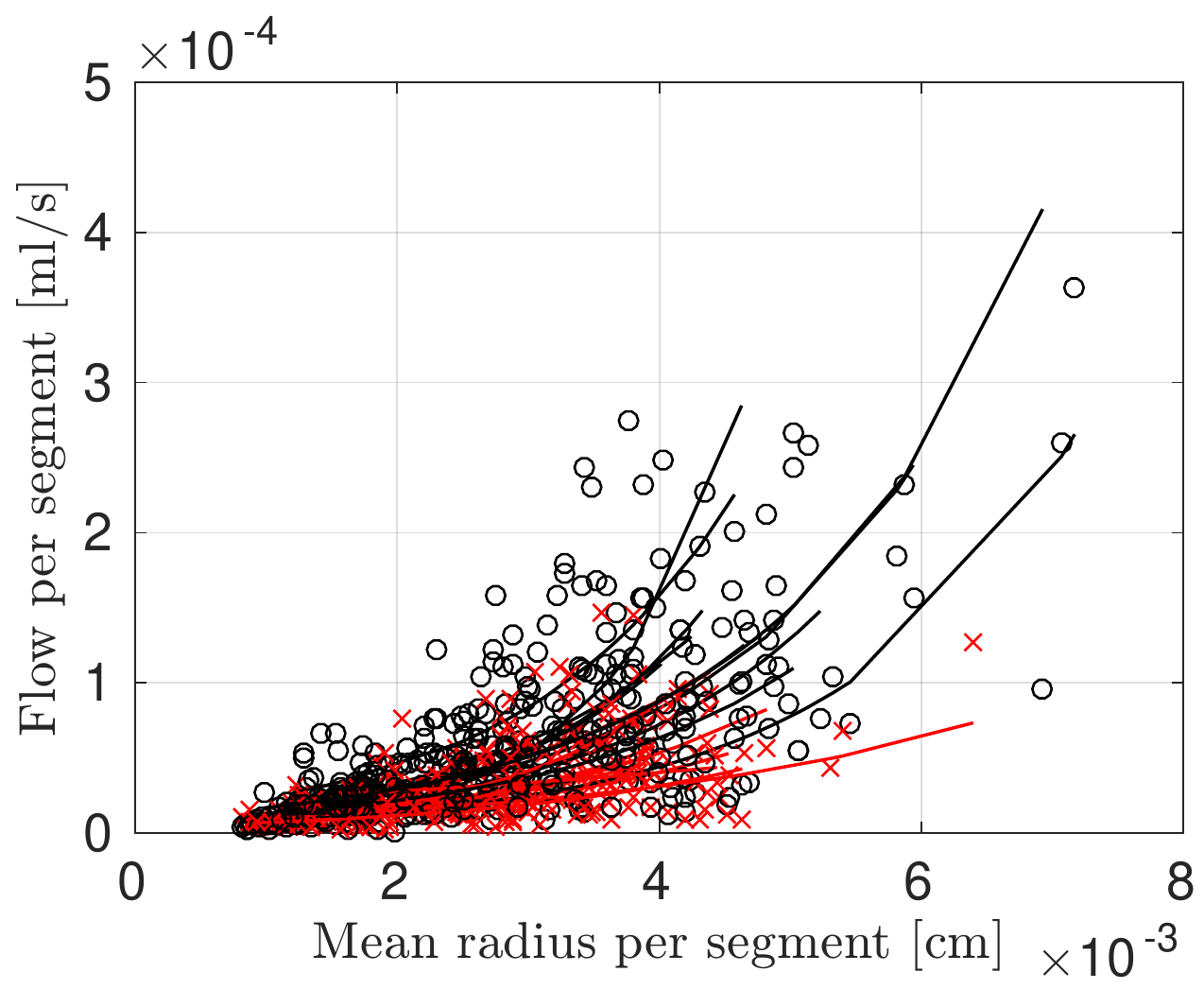}\label{fig:samples}}
  ~~
  \subfigure[Fitted curves]{\includegraphics[width=0.35\textwidth]{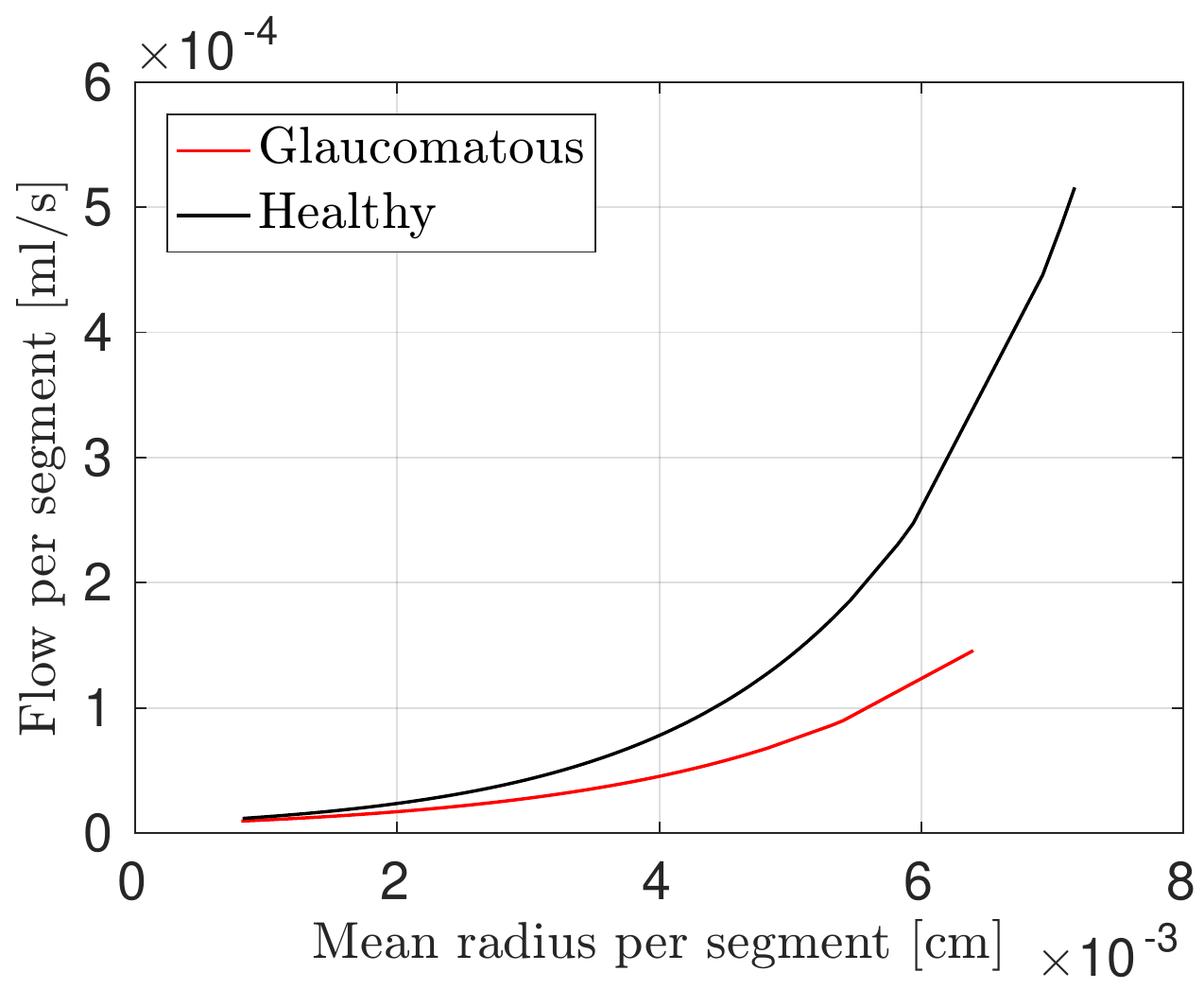}\label{fig:fitted}}
\caption{Mean radius vs. flow rate per segment for healthy (black) and glaucomatous (red) subjects in the LES-AV database. (a) Samples per image and fitted curves for each individual. (b) Fitted curves for each group. The SC1 and SC3 were used for the glaucomatous and healthy groups, respectively.}
\label{fig:fitted-curves}
\end{figure}

The hemodynamic quantities obtained using the outcomes of SC1 and SC3 for the G and H groups, respectively, are in line with clinical observations. Fig.~\ref{fig:samples} shows the flow vs. mean radius per segment for all subjects. 
The exponential curves fitted for each group are depicted in Fig.~\ref{fig:fitted}. The variables are correlated (Spearman's $\rho=0.71$, $p\ll0.01$), which is consistent with the literature~\cite{Liu_RCFD_FirstWork-2009}.

\input{results-tables.tex}

Finally, statistical analysis at a confidence level of 99\% were performed on a per-patient and a per-measurement basis. 
On the former (Table~\ref{TAB::Results::Statistics::PerPatient}), we focused on the age, sex and number of segments (NoS). The differences in the mean values of the NoS and the age were not significant (Wilcoxon test, $p>0.1$). Moreover, the patient diagnosis was independent from the sex ($\chi^2$ test, $p=1$).


On a per-measurement basis (Table~\ref{TAB::Results::Statistics::PerMeasurement}), all the arterial segments, terminal and bifurcations points were considered as separate measurements. 
The analysis was focused on the $\Delta P$, $v$, $r$, age and sex. 
The differences in the mean value of $\Delta P$ and $v$ were larger in the H group, 
while the age was larger in the G group ($p\ll0.01$) (two tailed Wilcoxon test). 
The $r$ was lower in the G group, although not significant ($p=0.31$).
As in the per-patient analysis, the diagnostic was independent from the sex ($\chi^2$ test, $p=0.02$). 
The Pearson correlation coefficient between age vs. $\Delta P$ and age vs. $v$ was close to zero and not significant, suggesting that the age is not an influential factor for the comparison of the means.



%% file: results-tables.tex
\begin{table*}[t]
    \begin{minipage}{.475\textwidth}
      \centering
      \resizebox{\textwidth}{!}{
      \begin{tabular}{| L{2cm} | C{2cm} | C{2cm} | C{2cm} |}
        \hline
        \textbf{Variable / Group}  & \textbf{All} ($n=22$) & \textbf{Healthy} ($n=11$)  & \textbf{Glaucomatous} ($n=11$)  \\
        \hline
        NoS           		   & 33.27$\pm$13.72 & 37.91$\pm$16.78 & 28.64$\pm$8.15\\
        Age \textit{[years]}   & 71.36$\pm$9.98	 & 68.64$\pm$9.24  & 74.09$\pm$10.36\\
        Sex \textit{(Males)}   & 12         	 & 6              	 & 6\\
        \hline
      \end{tabular}}
      \label{TAB::Results::Statistics::PerPatient}
      \caption{Summary of statistics at a per-patient level. None of the variables present statistically significant differences between groups ($p\gg0.01$).}
    \end{minipage}%
	\hspace{0.05\textwidth}
	\begin{minipage}{.475\linewidth}
      \centering
      \resizebox{\textwidth}{!}{
      \begin{tabular}{| L{2cm} | C{2cm} | C{2cm} | C{2cm} |}
      \hline
      \textbf{Variable / Group} 	   & \textbf{All} ($n=1466$) & \textbf{Healthy} ($n=836$) & \textbf{Glaucomatous} ($n=630$)  \\
      \hline
      $\Delta P$ [\textit{mmHg}]$^*$  & 6.94$\pm$7.40	& 8.08$\pm$8.12	  & 5.43$\pm$5.99 \\
      $v$ [\textit{cm/s}]       $^*$  & 2.26$\pm$1.84	& 2.61$\pm$2.03	  & 1.79$\pm$1.41 \\
      $r$ [\textit{cm}]               & 0.003$\pm$0.001	& 0.003$\pm$0.001 & 0.003$\pm$0.001 \\
      Age [\textit{years}]      $^*$  & 72.08$\pm$9.68	& 70.83$\pm$9.74  & 73.74$\pm$9.35 \\
      Sex \textit{(Males)}      $^\dagger$  & 763	&	413	&	350	\\
      \hline
      \end{tabular}
      }
      \label{TAB::Results::Statistics::PerMeasurement}
      \caption{Summary of statistics at a per-measurement level (terminals, bifurcations and arterial segments). ($^*$) points statistically significant differences between groups ($p\ll0.01$). ($^\dagger$) reached $p=0.02$.}
    \end{minipage} 
\end{table*}

%% file: discussion.tex
\section{Discussion}
\label{sec:discussion}

We have presented a first approach to characterize the retinal hemodynamics of glaucoma patients using computational fluid-dynamics. To the best of our knowledge, this is the first study in which the relationship between glaucoma and simulated hemodynamic outcomes of the retinal arterioles are studied. 

The arterial radius was previously observed to be smaller in glaucoma patients than in normal subjects~\cite{mitchell2005retinal}, a setting that is consistent with our data. On the other hand, in-vivo velocity measurements in the CRA of affected patients was shown to be lower than in control subjects~\cite{abegao2012disturbed}. Under these considerations, we have used SC1 for the G group and SC3 for the H group in our first pilot study, resulting in a simulated behavior that is in line with existing observations made on other measurable vessels (i.e. $\text{G}(v)<\text{H}(v)$)~\cite{abegao2012disturbed}. Moreover, this rendered physiologically correct values of $\Delta p$ in the retinal network.

The classification experiment based on BoHF and the unbiased SC2 outcomes showed an AUC different than random for identifying the glaucomatous patients, indicating that the tool is able to preserve the variations in the hemodynamics of the arteries affected by the disease. Future efforts will be focused on incorporating the venous network and also modeling of the IOP, which is relevant for clinical studies~\cite{abegao2013lack}. Also, the incorporation of a deep learning based method for simultaneous segmentation and classification of the vasculature will be explored in order to allow a more efficient characterization of larger populations.

%
%
%
%

Finally, our MATLAB/C++/python code and the LES-AV database are publicly released.
To the best of our knowledge, our data set is the first in providing not only the segmentations of the arterio-venous structures but also diagnostics and clinical parameters at an image level.
%
%
%
%

%% file: main.bbl
\begin{thebibliography}{10}

\bibitem{tham2014global}
Tham, Y.C.,  et~al.:
\newblock Global prevalence of glaucoma and projections of glaucoma burden
  through 2040: a systematic review and meta-analysis.
\newblock Ophthalmology \textbf{121}(11) (2014)  2081--2090

\bibitem{Harris_ImportanceOfMathematicalModelInGlaucoma-2013}
Harris, A.,  et~al.:
\newblock Ocular hemodynamics and glaucoma: the role of mathematical modeling.
\newblock Eur J Ophthalmol  139--146

\bibitem{barbosa2018clinical}
Barbosa-Breda, J.,  et~al.:
\newblock Clinical metabolomics and glaucoma.
\newblock Ophthalmic research \textbf{59}(1) (2018)  1--6

\bibitem{abegao2016ocular}
Abeg{\~a}o~Pinto, L.,  et~al.:
\newblock Ocular blood flow in glaucoma--the {Leuven Eye Study}.
\newblock Acta Ophthalmologica \textbf{94}(6) (2016)  592--598

\bibitem{Lu_RCFD_DiabeticParafovealStudy-2016}
Lu, Y.,  et~al.:
\newblock Computational fluid dynamics assisted characterization of parafoveal
  hemodynamics in normal and diabetic eyes using adaptive optics scanning laser
  ophthalmoscopy.
\newblock Biomed Opt Express \textbf{7}(12)  4958

\bibitem{Liu_RCFD_FirstWork-2009}
Liu, D.,  et~al.:
\newblock Image-based blood flow simulation in the retinal circulation.
\newblock In: IFMBE Proceedings, Springer (2009)  1963--1966

\bibitem{Ganesan_RCFD_GraphArteryVeinsRats2-2010}
Ganesan, P., He, S., Xu, H.:
\newblock Analysis of retinal circulation using an image-based network model of
  retinal vasculature.
\newblock Microvascular Research \textbf{80}(1)  99--109

\bibitem{Caliva_RCFD_DiabeticRetinopatyProgression-2017}
Caliva, F.,  et~al.:
\newblock Hemodynamics in the retinal vasculature during the progression of
  diabetic retinopathy.
\newblock JMO \textbf{1}(4) (2017)  6--15

\bibitem{fei2005bayesian}
Fei-Fei, L., Perona, P.:
\newblock A bayesian hierarchical model for learning natural scene categories.
\newblock In: CVPR. Volume~2., IEEE (2005)  524--531

\bibitem{moccia2018blood}
Moccia, S.,  et~al.:
\newblock Blood vessel segmentation algorithms—review of methods, datasets
  and evaluation metrics.
\newblock CMPB \textbf{158} (2018)  71--91

\bibitem{giancardo2017representation}
Giancardo, L., Roberts, K., Zhao, Z.:
\newblock Representation learning for retinal vasculature embeddings.
\newblock In: Fetal, Infant and Ophthalmic Medical Image Analysis.
\newblock Springer (2017)  243--250

\bibitem{rumpf2002continuous}
Rumpf, M., Telea, A.:
\newblock A continuous skeletonization method based on level sets.
\newblock In: Eurographics / IEEE VGTC Symposium on Visualization. (2002)
  151--159

\bibitem{Maurer_EuclideanTransformUsedByMATLAB-2003}
Maurer, C.R., Qi, R., Raghavan, V.:
\newblock A linear time algorithm for computing exact {Euclidean} distance
  transforms of binary images in arbitrary dimensions.
\newblock IEEE PAMI \textbf{25}(2) (2003)  265--270

\bibitem{Pries_VariablViscosity-1996}
Pries, A.R., Secomb, T.W., Gaehtgens, P.:
\newblock Biophysical aspects of blood flow in the microvasculature.
\newblock Cardiovasc Res \textbf{32}(4)  654--667

\bibitem{Blanco_CCO-2013}
Blanco, P., Queiroz, R., Feijóo, R.:
\newblock A computational approach to generate concurrent arterial networks in
  vascular territories.
\newblock Int J Numer Method Biomed Eng \textbf{29} (2013)  601--614

\bibitem{pournaras2013retinal}
Pournaras, C.J., Riva, C.E.:
\newblock Retinal blood flow evaluation.
\newblock Ophthalmologica \textbf{229}(2) (2013)  61--74

\bibitem{mitchell2005retinal}
Mitchell, P.,  et~al.:
\newblock Retinal vessel diameter and open-angle glaucoma: the {Blue Mountains
  Eye Study}.
\newblock Ophthalmology \textbf{112}(2) (2005)  245--250

\bibitem{abegao2012disturbed}
Abeg{\~a}o~Pinto, L., Vandewalle, E., Stalmans, I.:
\newblock Disturbed correlation between arterial resistance and pulsatility in
  glaucoma patients.
\newblock Acta Ophthalmol \textbf{90}(3) (2012)

\bibitem{abegao2013lack}
Abeg{\~a}o~Pinto, L.,  et~al.:
\newblock Lack of spontaneous venous pulsation: possible risk indicator in
  normal tension glaucoma?
\newblock Acta Ophthalmol \textbf{91}(6) (2013)  514--520

\end{thebibliography}
